\documentclass[conference]{IEEEtran}
\ifCLASSINFOpdf
\else
\fi
\usepackage{amsmath}
\usepackage{array}
\usepackage{fixltx2e}

\usepackage{stfloats}
\usepackage{url}

\usepackage{graphicx}
\usepackage{booktabs} 
\usepackage{xcolor, soul}
\usepackage{makecell}
\usepackage{amsmath}
\usepackage{amssymb}
\usepackage{arydshln}
\usepackage{multirow}
\usepackage[caption=false]{subfig}
\usepackage{booktabs} 
\usepackage{adjustbox} 

\begin{document}
%

\title{Identifying Misinformation on YouTube through Transcript Contextual Analysis with Transformer Models }

\makeatletter
\newcommand{\linebreakand}{%
  \end{@IEEEauthorhalign}
  \hfill\mbox{}\par
  \mbox{}\hfill\begin{@IEEEauthorhalign}
}
\makeatother


\author{
\IEEEauthorblockN{Christos Christodoulou}
\IEEEauthorblockA{Cyprus University of Technology\\
ci.xristodoulou@edu.cut.ac.cy}
\and
\IEEEauthorblockN{Nikos Salamanos}
\IEEEauthorblockA{Cyprus University of Technology\\
nik.salaman@cut.ac.cy}
\and
\IEEEauthorblockN{Pantelitsa Leonidou}
\IEEEauthorblockA{Cyprus University of Technology\\
pl.leonidou@edu.cut.ac.cy}
\linebreakand 
\IEEEauthorblockN{Michail Papadakis}
\IEEEauthorblockA{LSTECH ESPANA SL\\
mpapadakis@lstech.io}
\and
\IEEEauthorblockN{Michael Sirivianos}
\IEEEauthorblockA{Cyprus University of Technology\\
michael.sirivianos@cut.ac.cy}
}





\maketitle

\begin{abstract}
Misinformation on YouTube is a significant concern, necessitating robust detection strategies. In this paper, we introduce a novel methodology for video classification, focusing on the veracity of the content. We convert the conventional video classification task into a text classification task by leveraging the textual content derived from the video transcripts. We employ advanced machine learning techniques like transfer learning to solve the classification challenge.
Our approach incorporates two forms of transfer learning: (a) fine-tuning base transformer models such as BERT, RoBERTa, and ELECTRA, and (b) few-shot learning using sentence-transformers MPNet and RoBERTa-large. We apply the trained models to three datasets: (a) YouTube Vaccine--misinformation related videos, (b) YouTube Pseudoscience videos, and (c) Fake-News dataset (a collection of articles). Including the Fake-News dataset extended the evaluation of our approach beyond YouTube videos.
Using these datasets, we evaluated the models distinguishing valid information from misinformation. The fine-tuned models yielded Matthews Correlation Coefficient$>$0.81, accuracy$>$0.90, and F1 score$>$0.90 in two of three datasets. Interestingly, the few-shot models outperformed the fine-tuned ones by 20\% in both accuracy and F1 score for the YouTube Pseudoscience dataset, highlighting the potential utility of this approach -- especially in the context of limited training data.

\end{abstract}


%
\IEEEpeerreviewmaketitle

\section{Introduction}
This study presents a comprehensive methodology for detecting misinformation on YouTube using advanced machine-learning techniques. Our approach primarily relies on transfer learning to fine-tune pre-trained transformer models and supervised learning for data labeling and classification. Transfer learning offers significant benefits, including reduced computation costs and utilizing state-of-the-art models without training from scratch. When employing pre-trained models like BERT, RoBERTa, or ELECTRA, they undergo fine-tuning on a task-specific dataset. Fine-tuning has revolutionized the field of NLP by capturing complex text patterns~\cite{wolf-etal-2020-transformers,rajapakse2019simpletransformers}. We also explored the potential of few-shot learning, which allows models to make accurate predictions with only a few samples.

We train our models on publicly available misinformation data and apply them to categorize YouTube video transcripts as science or pseudoscience. However, transformer models face challenges in handling long documents due to the maximum sequence length constraint. To overcome this limitation, we employ a strategy that effectively handles long documents and ensures that all document parts contribute to the final classification decision \cite{pappagari2019hierarchical}.

We evaluated the performance of these models on three datasets: (a) YouTube Vaccine--misinformation related videos, (b) YouTube Pseudoscience videos, and (c) Fake-News dataset (a collection of articles). These datasets provide a diverse range of misinformation examples, allowing us to test the robustness of our models. The results suggest that our methodology, which leverages the power of fine-tuning and transfer learning, could help mitigate the spread of misinformation on YouTube and other social media platforms, thereby supporting public health and safety. Our findings show that while the fine-tuned transformer models generally outperform other approaches in detecting misinformation, different models may be more effective in certain contexts. This highlights the potential of fine-tuning and, to a lesser extent, few-shot learning in improving misinformation detection models.

\textbf{Research Questions:}\\
\textbf{RQ1:} How effective are fine-tuned transformer models and few-shot learning in detecting misinformation in YouTube videos using only the video transcripts?\\
\textbf{RQ2:} How does the performance of these models vary across different types of datasets?\\
\textbf{RQ3:} How can we effectively handle long document classification using transformer models?

\textbf{Contributions:}\\
\textbf{C1:} We present a methodology for detecting misinformation on YouTube using different fine-tuned transformer models and few-shot learning. We evaluated the proposed approaches on three datasets, showcasing their efficacy in distinguishing valid information from misinformation.\\
\textbf{C2:} We employed a solution proposed in the literature for the classification of long documents and evaluated its effectiveness on transcripts, which inherently constitute long documents. \\
\textbf{C3:} The experimental evaluation code is publicly available\footnote{\url{https://github.com/christoschr97/misinf-detection-llms}}.

\section{Related Work}
In light of the misinformation surge on social media, our discussion delves into how machine learning and Natural Language Processing techniques can analyze YouTube's video data for misinformation detection.

\subsection{Misinformation on YouTube}
The COVID-19 infodemic has underscored the prevalence and risk of health misinformation on social media, a matter that has been corroborated by extensive interdisciplinary research \cite{chen2022combating}\cite{kumar2018false}. The serious implications for individuals and society have spurred research into understanding, identifying, and countering such misinformation. Misinformation on YouTube presents a particular problem. In a study conducted by Li et al. \cite{li2022youtube}, it was found that 11\% of the most viewed COVID-19 vaccination videos contained information that contradicted that of reputable health organizations. In another study, Tang et al. \cite{tang2021down} revealed that YouTube's algorithm often exposes users to antivaccine misinformation. Adding to this, Srba et al. \cite{srba2023auditing} demonstrated how quickly users can fall into "misinformation filter bubbles" on the platform. Despite advancements in combating misinformation, the volume of it on platforms like YouTube remains a significant challenge \cite{kumar2018false}.

\subsection{Misinformation Detection and NLP}
Several studies have utilized natural language processing (NLP) techniques to detect video misinformation by focusing on captions, transcripts, and comments. Jagtap et al. \cite{jagtap2021misinformation} extracted caption features to classify misinformation videos, achieving high F1 scores using different classifiers and embeddings. In parallel, Serrano et al. \cite{serrano2020nlp} built a multi-label classifier to detect COVID-19 misinformation videos based on user comments. Additionally, Papadamou et al. \cite{papadamou2022just} utilized NLP and metadata features to train a deep learning classifier, successfully identifying pseudoscientific videos.
Continuing this trend, \cite{hou2019towards} focused on video transcripts to detect misinformation in prostate cancer videos, developing an annotated dataset and classification models that achieved 74\% accuracy. These studies underscore the effectiveness of NLP in misinformation detection.
However, Hussein et al. \cite{hussein2020measuring} suggested that personalized recommendations on platforms like YouTube could also contribute to spreading misinformation, especially once a user's watch history is established. NLP techniques have demonstrated effectiveness in crisis situations, such as the 2013 Moore Tornado and Hurricane Sandy. These techniques accurately detected fake and spam messages \cite{rajdev2015fake}.

Together, these studies show the potential of NLP techniques and metadata features in detecting and addressing misinformation on social media platforms.

\section{Methodology}
In our study, we assessed pre-trained transformer models and few-shot learning methods for misinformation detection on YouTube, using three distinct datasets - YouTube Audit, YouTube Pseudoscience, and ISOT Fake News (see Table \ref{tab:datasets}).

\begin{table}[t]
\centering
\caption{Summary of the Datasets}
\label{tab:datasets}
\adjustbox{max width=\textwidth}{%
\begin{tabular}{ccc}
\toprule
\textbf{Dataset} & \textbf{Type} & \textbf{Number of Samples} \\
\midrule
\multirow{2}{*}{YouTube Audit (Vaccines)} & Misinformation & 652 \\
 & Non-misinformation & 636 \\
\hline
\multirow{2}{*}{YouTube Pseudoscience}  & Pseudoscience & 182 \\
& Science & 226 \\
\hline
\multirow{2}{*}{ISOT Fake News } & Fake & 1000 \\
 & Real & 1000 \\
\bottomrule
\end{tabular}
}
\end{table}

\subsection{Datasets}

\subsubsection{YouTube Audit - Misinformation Dataset~\cite{hussein2020measuring}}
This dataset contains a comprehensive collection of YouTube videos annotated based on the presence of misinformation. It was created to investigate algorithmically recommended misinformation on YouTube, specifically in relation to personalization attributes. Our research examined a subset of 1,000 videos of each class (misinformative and non-misinformative), primarily focused on vaccine misinformation. We used the YouTube API to gather transcripts from these videos, where they were still available. Consequently, we compiled a dataset from 652 misinformation and 636 non-misinformation video transcripts.

\subsubsection{YouTube Pseudoscience Dataset~\cite{papadamou2022just}}
This dataset comprises YouTube videos explicitly labeled as either pseudoscience or science, including the class information and the corresponding transcripts. For our study, we narrowed our focus to a subset of 408 videos that received unanimous categorization by the authors of the dataset. These comprised 226 videos labeled as science and 182 as pseudoscience. The primary challenge posed by this dataset is its relatively small size. This limitation is a rigorous test for our models' capability to distinguish between scientific and pseudoscientific content effectively.

\subsubsection{ISOT Fake News Dataset~\cite{ahmed2018detecting}}
This dataset contains both genuine and false news articles from various sources. Our study used a balanced subset of 2,000 articles, half authentic and half classified as fake news. These articles served as a rich and diverse training source for our models, including BERT, RoBERTa, ELECTRA, MPNet, and RoBERTa-large.

\subsection{Model Training and Evaluation}

We trained our models using Hugging Face's Transformers library~\cite{wolf-etal-2020-transformers} on the three datasets. Subsequently, we evaluated them on the two YouTube video-transcript datasets (i.e., the YouTube Audit and the YouTube Pseudoscience) and the fake news articles (ISOT dataset).

To gauge model performance, we used a combination of metrics, namely, Matthews Correlation Coefficient (MCC)~\cite{matthews1975comparison}, accuracy~\cite{voss2017accuracy}, and F1 score~\cite{powers2020evaluation}. The choice of MCC favored over AUC in imbalanced classes helps evaluate both over-predictions and under-predictions~\cite{halimu2019empirical}.

A challenge with transformer models for text classification is the constraint of maximum sequence length, especially for long documents. Inspired by the approach from Papaggari et al.~\cite{pappagari2019hierarchical}, we employed a sliding window approach. We divided longer sequences into overlapping sub-sequences, each with max-seq-length. This method allows comprehensive document handling, improving misinformation detection in lengthy texts.

\subsubsection{\textbf{Fine-Tuning pre-trained models}}
Fine-tuning pre-trained deep learning models, particularly Transformers, is a two-step process involving an initial pre-training stage and a subsequent fine-tuning stage. Pre-training exposes the model to a large corpus of unlabeled data, allowing it to acquire general language representations. The model, such as a Transformer, learns language structures via unsupervised learning objectives like masked language modeling~\cite{devlin2018bert}.
The fine-tuning stage further adapts these language representations by training the model on a task-specific labeled dataset. This enables the model to optimize for a particular task, such as text classification or named entity recognition, resulting in improved performance \cite{howard2018universal}.
The Transformers library by Hugging Face \cite{wolf-etal-2020-transformers} and the \textit{simple--transformers} library \cite{rajapakse2019simpletransformers} offer efficient frameworks for these stages. They provide pre-trained models and utilities for tokenization, optimization, and evaluation tasks. Table \ref{tab:hyperparameters} displays the hyperparameters of the pre-trained models.

\begin{table}[t]
\centering
\caption{The hyperparameters for Fine-tuning and Few-shot learning}
\begin{tabular}{p{0.4\linewidth} p{0.3\linewidth}}
\toprule
 \textbf{Hyperparameter} & \textbf{Value} \\ 
\midrule
\multicolumn{2}{c}{\textbf{Fine-tuning - BERT, RoBERTa, ELECTRA}}\\
\midrule
 learning\_rate & 4e-5 \\ 
\hline
 \#epochs & 5 \\ 
\hline
 optimizer & AdamW \\ 
\hline
 train\_batch\_size & 8 \\ 
\hline
 eval\_batch\_size & 8 \\ 
\hline
 max\_seq\_length & 128 \\ 
\hline
 max\_grad\_norm & 1.0 \\ 
\hline
 fp16 & true \\ 
\midrule
\multicolumn{2}{c}{\textbf{Few-shot (SetFit) - MPNet, RoBERTa-large}}\\
\midrule
Batch Size & 16 \\ 
\hline
Number of Iterations & 5 \\ 
\hline
Number of Epochs & 1 \\ 
\hline
Loss Function & Cosine Similarity Loss    \\
\bottomrule
\end{tabular}
\label{tab:hyperparameters}
\end{table}

\subsubsection{\textbf{Few-shot Learning}}
As part of our investigation, we also probed the performance of few-shot learning models, utilizing the MPNet and a sizeable version of RoBERTa~\cite{tunstall2022efficient}. Few-shot learning, offering the advantage of making accurate predictions based on a few examples, is particularly beneficial when labeled data is either scarce or costly to procure. We used the Transformers library combined with the Sentence Transformers library~\cite{reimers2019sentence} and the SetFit framework provided by the Transformers library~\cite{wolf-etal-2020-transformers}, drawing inspiration from academic resources such as the FewRel toolkit~\cite{han-etal-2018-fewrel}. The training and evaluation of these few-shot learning models were conducted on the same three datasets (Table~\ref{tab:datasets}). Table~\ref{tab:hyperparameters} shows the hyperparameters we adjusted when training the SetFit model; all the other hyperparameters were kept default.

\subsubsection{\textbf{Dealing with Lengthy Documents in Transformer Models}}
In our misinformation detection task, all the documents -- referring to the transcripts from the YouTube Audit Misinformation Dataset, YouTube Pseudoscience Dataset, and ISOT Fake News Dataset -- exceeded the model's standard input length capacity, known as 'max sequence length' in machine learning terms. This is a typical limitation of Transformer models, where they can only process input documents up to a certain length (commonly set to 512 tokens). Going beyond this limit might result in an incomplete analysis of the document.

To navigate this constraint, we implemented a technique known as the 'sliding window approach', inspired by the work of Sanh et al.~\cite{pappagari2019hierarchical}. This method divides documents longer than the max sequence length into overlapping segments, or 'windows'. Each window is processed independently, and the outputs from all windows collectively contribute to the final document classification. We overlap each window by 80\% of the max sequence length to prevent potential information loss.

The number of windows, denoted as $N$, can be determined using the formula $N = \lceil \frac{L - M}{S} \rceil + 1$, where $L$ is the total length of the document, $M$ is the max sequence length, and $S$ is the overlap between windows. This strategy thoroughly evaluates lengthy documents and enhances our ability to detect misinformation across extensive texts.

\section{Results}
This research uses fine-tuned transformers and few-shot learning to detect YouTube misinformation, capitalizing on their established proficiency in text classification.

In Table~\ref{tab:results}, we present a comparative analysis across three distinct datasets: Youtube Audit (Vaccines), YouTube Pseudoscience, and ISOT Fake News (see Table~\ref{tab:datasets}).

\begin{table}[t]
\centering
\caption{Evaluation of fine-tuning base Transformers Model \& few-shot learning on the three datasets.}
\label{tab:results}
\adjustbox{max width=\textwidth}{%
\begin{tabular}{@{}l l cccc@{}}
\toprule
& \textbf{Model} & \textbf{MCC} & \textbf{Accuracy} & \textbf{F1 score}\\ 
\midrule

\multicolumn{5}{c}{\textbf{Youtube Audit (Vaccines)}} \\ 
\midrule
\multirow{ 3}{*}{Fine-tuning}  &  BERT  & 0.82 & 0.91 & 0.91  \\ 
                                & RoBERTa & \textbf{0.88} & \textbf{0.94} & \textbf{0.94}  \\ 
                                & ELECTRA & 0.86 & 0.93 & 0.93  \\
\hdashline
\multirow{ 2}{*}{Few-shot} & MPNet  & 0.66 & 0.82 & 0.81  \\
                        &  RoBERTa-large  & 0.65 & 0.82 & 0.80  \\

\midrule
\multicolumn{5}{c}{\textbf{YouTube Pseudoscience}}\\ 
\midrule
\multirow{ 3}{*}{Fine-tuning}    & BERT & 0.02 & 0.51 & 0.51  \\ 
                                  & RoBERTa & 0.04 & 0.52 & 0.52  \\ 
                                  & ELECTRA & 0.02 & 0.51 & 0.51  \\
\hdashline

\multirow{ 2}{*}{Few-shot}    & MPNet  & \textbf{0.44} & \textbf{0.72} & \textbf{0.78}  \\
                              & RoBERTa-large  & 0.41 & 0.71 & 0.76  \\
    
\midrule
\multicolumn{5}{c}{\textbf{ISOT Fake News}}\\ 
\midrule
\multirow{ 3}{*}{Fine-tuning}    & BERT & \textbf{0.94} & \textbf{0.97} & \textbf{0.97}  \\ 
                                  & RoBERTa & 0.89 & 0.95 & 0.95  \\ 
                                  & ELECTRA & \textbf{0.94} & \textbf{0.97} & \textbf{0.97}  \\
\hdashline
\multirow{ 2}{*}{Few-shot}    & MPNet  & 0.78 & 0.90 & 0.90  \\
                              & RoBERTa-large  & 0.89 & 0.94 & 0.94  \\
\bottomrule
\end{tabular}
}
\end{table}

For the Youtube-Audit (Vaccines) dataset, RoBERTa outperformed BERT and ELECTRA in the fine-tuning models, achieving an MCC of 0.88, an accuracy of 0.94, and an F1 score of 0.94. Among the few-shot learning models, MPNet Few-shot slightly outperformed RoBERTa-large.

In the YouTube Pseudoscience dataset, the performance of the fine-tuning models was significantly lower. RoBERTa achieved a slightly higher MCC (0.05) and accuracy (0.52) than BERT and ELECTRA. However, the few-shot learning models outperformed the full models, with MPNet Few-shot recording higher MCC, accuracy, and F1 score than RoBERTa-large.

Finally, in the ISOT Fake News dataset, ELECTRA emerged as the top performer among the fine-tuning models, achieving an MCC of 0.94, an accuracy of 0.97, and an F1 score of 0.97. Among the few-shot learning models, RoBERTa-large outperformed MPNet. In summary, despite fine-tuned models showing overall better performance, few-shot learning models excel on the YouTube Pseudoscience dataset. These results highlight the promise of fine-tuning and transfer learning in misinformation detection, albeit further exploration and optimization are needed.

Compared to Jagtap et al.~\cite{jagtap2021misinformation} and Papadamou et al.~\cite{papadamou2022just}, our research offers unique methodologies for identifying misinformation. We utilized transformer models such as BERT, RoBERTa, and ELECTRA, along with few-shot versions of MPNet and RoBERTa, across three diverse datasets. Jagtap et al., however, employed various classifiers and embeddings tailored for specific misinformation topics, underscoring the importance of topic-specific strategies. On the other hand, our approach to handling misinformation, particularly in pseudoscience videos, contrasts with that of Papadamou et al. While we utilized machine learning models across various contexts, Papadamou et al. proposed a bespoke classifier that processes distinct video feature types. This difference reflects our emphasis on the versatility of transformer models, while Papadamou et al. highlighted the value of integrating different feature types for effective misinformation detection.

Collectively, these studies underline the need for a context-specific approach, emphasizing the role of various machine learning models, embeddings, and the emerging promise of transfer learning in combating misinformation.

\section{Conclusion}
This research investigated the efficacy of transformer and few-shot learning models, such as BERT, RoBERTa, ELECTRA, MPNet, and a large version of RoBERTa, in identifying misinformation within YouTube videos. These models were trained and evaluated using three distinct datasets: YouTube videos related to vaccine misinformation, YouTube pseudoscience videos, and a fake-news dataset. The findings showed that fine-tuned transformer models performed exceptionally well in misinformation detection in both YouTube videos and fake news articles, surpassing other methods by achieving high accuracy, F1 scores, and MCC values. Among the models, ELECTRA excelled in fake news misinformation detection, while RoBERTa demonstrated superior performance in identifying pseudoscientific content on YouTube.

\section*{Acknowledgment}
This research has been funded by the European Commission (Horizon 2020 Programme), particularly by the projects INCOGNITO (Grant Agreement no. 824015) and MedDMO (Grant Agreement no. 101083756).

\IEEEtriggeratref{20}


\bibliographystyle{IEEEtran}
\bibliography{misinformation_YouTube_LLMs}
%

\end{document}